\pdfoutput=1

\documentclass[11pt]{article}
\usepackage[table]{xcolor}

\usepackage[]{acl}
\usepackage{times}
\usepackage{latexsym}

\usepackage[T1]{fontenc}
\usepackage[utf8]{inputenc}

\usepackage{microtype}

\usepackage{inconsolata}

\usepackage{graphicx}

\usepackage{hyperref}
\usepackage{url}
\usepackage{booktabs} 
\usepackage{graphicx}
\usepackage{algorithm}
\usepackage{algpseudocode}
\usepackage{xcolor}  
\usepackage{enumitem}
\usepackage{multirow}
\usepackage{tcolorbox}
\usepackage{colortbl}
\usepackage{pgf}
\usepackage{soul}
\usepackage{amsmath}
\usepackage{tcolorbox}
\usepackage{listings}
\usepackage{cleveref}

\newcommand{\method}[0]{\textsc{Prove}}

\definecolor{lightgreen}{rgb}{0.90, 0.97, 0.90}
\definecolor{mediumgreen}{rgb}{0.80, 0.90, 0.80}
\definecolor{darkgreen}{rgb}{0.70, 0.85, 0.70}

\newcommand{\highlight}[2]{\sethlcolor{#1}\hl{#2}}

\newcommand{\highlightvalue}[1]{%
    \begingroup
    \pgfmathsetmacro{\temp}{#1} 
    \ifdim \temp pt > 5pt
        \highlight{darkgreen}{(+#1)}
    \else\ifdim \temp pt > 2pt
        \highlight{mediumgreen}{(+#1)}
    \else\ifdim \temp pt > 0pt
        \highlight{lightgreen}{(+#1)}
    \else
        (+#1)
    \fi\fi\fi
    \endgroup
}

\title{Not All Votes Count! Programs as Verifiers Improve Self-Consistency of Language Models for Math Reasoning}

\author{Vernon Y.H. Toh$^{1}$, Deepanway Ghosal$^1$\thanks{Now at Deepmind.}, Soujanya Poria$^1$ \\\\
$^1$ Singapore University of Technology and Design
}

\definecolor{lightblue}{RGB}{220,230,241} %
\definecolor{lightblue2}{RGB}{180,210,230} %
\definecolor{lightblue3}{RGB}{140,190,220} %

\begin{document}
\maketitle

\begin{abstract}

Large language models (LLMs) have shown increasing competence in solving mathematical reasoning problems. However, many open-source LLMs still struggle with errors in calculation and semantic understanding during intermediate reasoning steps. In this work, we introduce \method{}, a simple yet effective framework that leverages translated programs derived from natural language solutions as a verification mechanism to filter out potentially incorrect reasoning paths before aggregating final answers. Unlike vanilla majority voting, our approach filters out solutions whose corresponding program output is inconsistent with the generated solution, aggregating only those that pass verification. We conducted extensive experiments using 13 open-source LLMs from various model families and sizes, ranging from 0.5B to 13B parameters, across eight mathematical benchmarks. Our results show that \method{} consistently outperforms vanilla majority voting as a heuristic for solving mathematical reasoning tasks across all model sizes and datasets, achieving improvements of up to 18\% on GSM8K and 8\% on MATH-500.  
Our codes are available at \url{https://github.com/declare-lab/prove}.
\end{abstract}

\begin{figure}[t]
    \centering
    \includegraphics[width=0.9\linewidth]{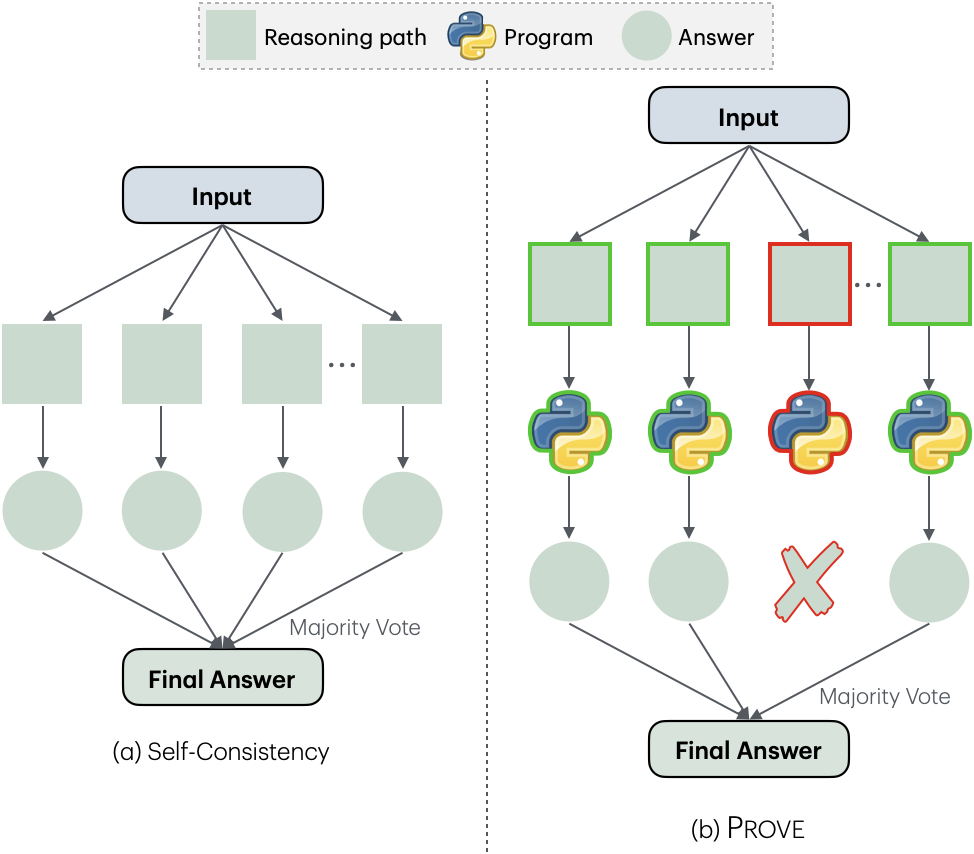}
    \caption{Comparison of self-consistency and \method{}. }
    \label{fig:framework_comparison}
\end{figure}

\section{Introduction}

Large Language Models (LLMs) have demonstrated remarkable success in mathematical reasoning tasks, especially with advanced models like GPT-4 \citep{openai2024gpt4technicalreport}. However, smaller open-source LLMs, such as Mistral, Gemma 2, and Llama 2 \cite{jiang2023mistral7b,gemmateam2024gemma2improvingopen,touvron2023llama2openfoundation} fall short in mathematical reasoning, often producing inaccurate content and failing to handle complex calculations. Many works since then have aimed to tackle those challenges by using different prompting methods such as Plan-and-Solve (PS) Prompting \citep{wang-etal-2023-plan} which aims to mitigate calculation errors and improve the overall quality of the generated reasoning by adopting a two-step process.
Alternatively, non-prompting approaches, such as fine-tuning LLMs specifically for reasoning tasks, have also shown notable improvements. However, this often demands large amounts of distilled data from models like GPT-4 \citep{wang2024mathcoder, gou2024tora}, resulting in significant computational and data generation costs.

\begin{figure*}[t]
    \centering
    \includegraphics[width=1.0\linewidth]{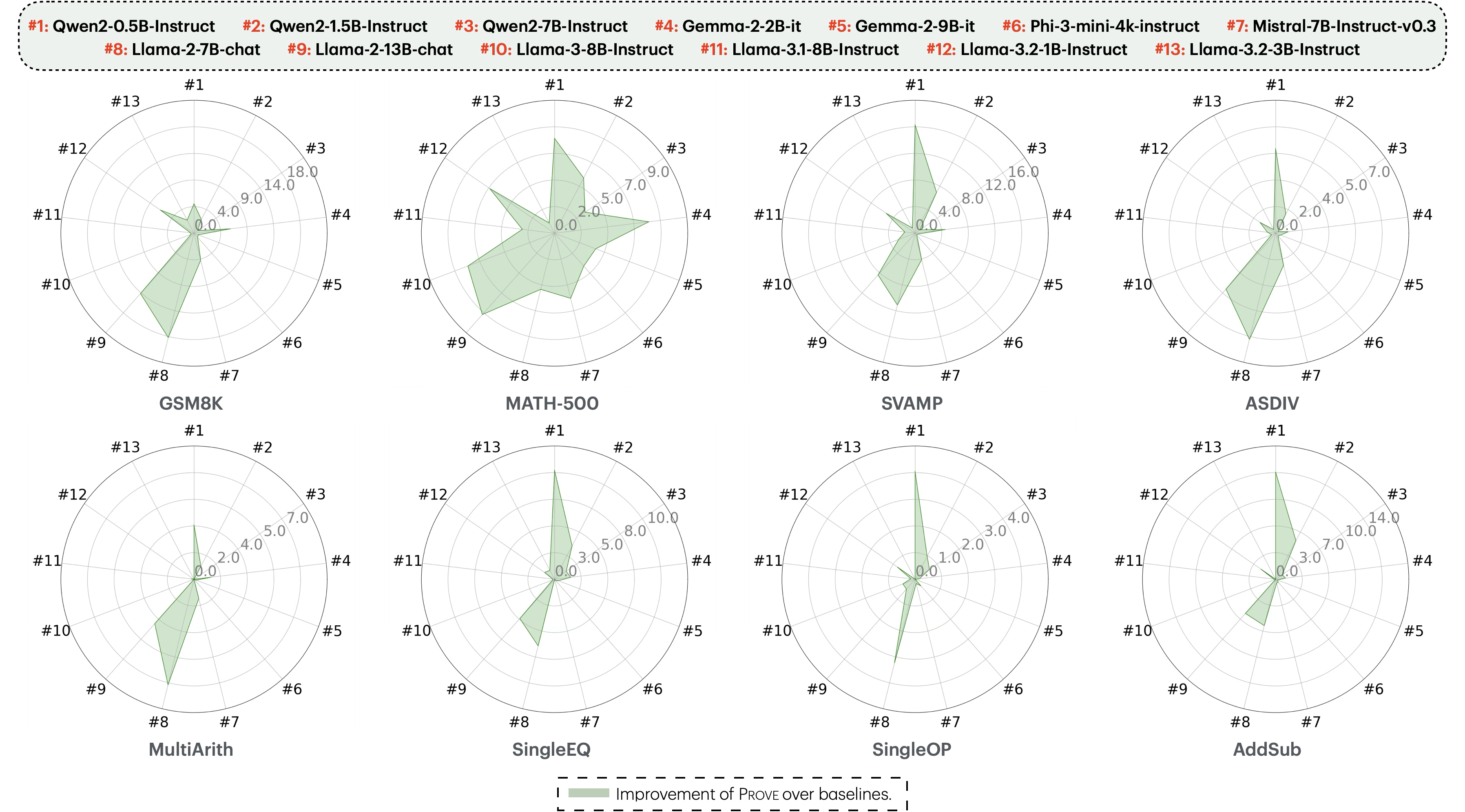}
    \caption{Performance improvement of \method{} over baseline methods across 13 LLMs evaluated on eight mathematical reasoning benchmarks.}
    \label{fig:framework_improvement}
\end{figure*}

Despite these advancements, fundamental challenges remain \citep{banerjee2024llmshallucinateneedlive}. Autoregressive models lack mechanisms to correct their own errors, meaning that once a mistake is made, it can propagate through the entire response. As a result, LLMs are still vulnerable to hallucinations which can lead to incorrect reasoning or calculation. Given these limitations, a promising direction is to enhance reasoning by leveraging the knowledge already embedded within LLMs themselves \citep{wang2023selfconsistency,hao-etal-2023-reasoning,NEURIPS2023_91edff07,weng-etal-2023-large,wang-etal-2023-plan}. Following \citet{wang2023selfconsistency}, one can sample multiple reasoning paths to solve a question and select the most common answer. While aggregating answers like this improves performance, sometimes the most common answer could still be wrong. 

Recent efforts have tried to develop better heuristics for selecting answers from multiple reasoning paths, such as verification via formalization \citep{lin-etal-2024-just}. However, translating problems and solutions into formal languages is often complex, requiring few-shot demonstration examples and more advanced models like GPT-3.5 to achieve effective verification. A simpler and more efficient alternative would be to directly leverage the code generation abilities of LLMs to do code-based verification without the overhead of formalization.

In this paper, we introduce \method{}, a \emph{frustratingly simple} approach that utilizes translated \textit{\textbf{PRO}grams for \textbf{VE}rification}. Our framework is model-agnostic and eliminates the need for any LLM fine-tuning or few-shot prompting. Instead, it leverages the capabilities of open-source LLMs to translate natural language plan and solution into Python program for verification. Unlike methods that rely on simple majority voting across all generated natural language solutions, we aggregate only those solutions that are successfully verified by their corresponding Python program. While a natural language plan and solution provides an abstract outline of the key steps in solving a math word problem, verifying its correctness can be challenging. Python program implementations offer a more precise representation, making it easier to accurately identify correct natural language solutions.

We evaluate \method{} using 13 LLMs from various model families and sizes \citep{abdin2024phi3technicalreporthighly,touvron2023llama2openfoundation,dubey2024llama3herdmodels,jiang2023mistral7b,gemmateam2024gemma2improvingopen,yang2024qwen2technicalreport}, ranging from 0.5B to 13B parameters across eight mathematical reasoning datasets \citep{cobbe2021gsm8k,patel-etal-2021-nlp,miao-etal-2020-diverse,roy-roth-2015-solving,koncel-kedziorski-etal-2015-parsing,roy-etal-2015-reasoning,hosseini-etal-2014-learning,lightman2024lets}. The results demonstrate that \method{} consistently outperforms vanilla majority voting \citep{wang2023selfconsistency} across all model sizes and datasets, with up to 18\% improvement on GSM8K and 8\% on MATH-500.

\begin{figure*}[ht]
    \centering
    \includegraphics[width=1\linewidth]{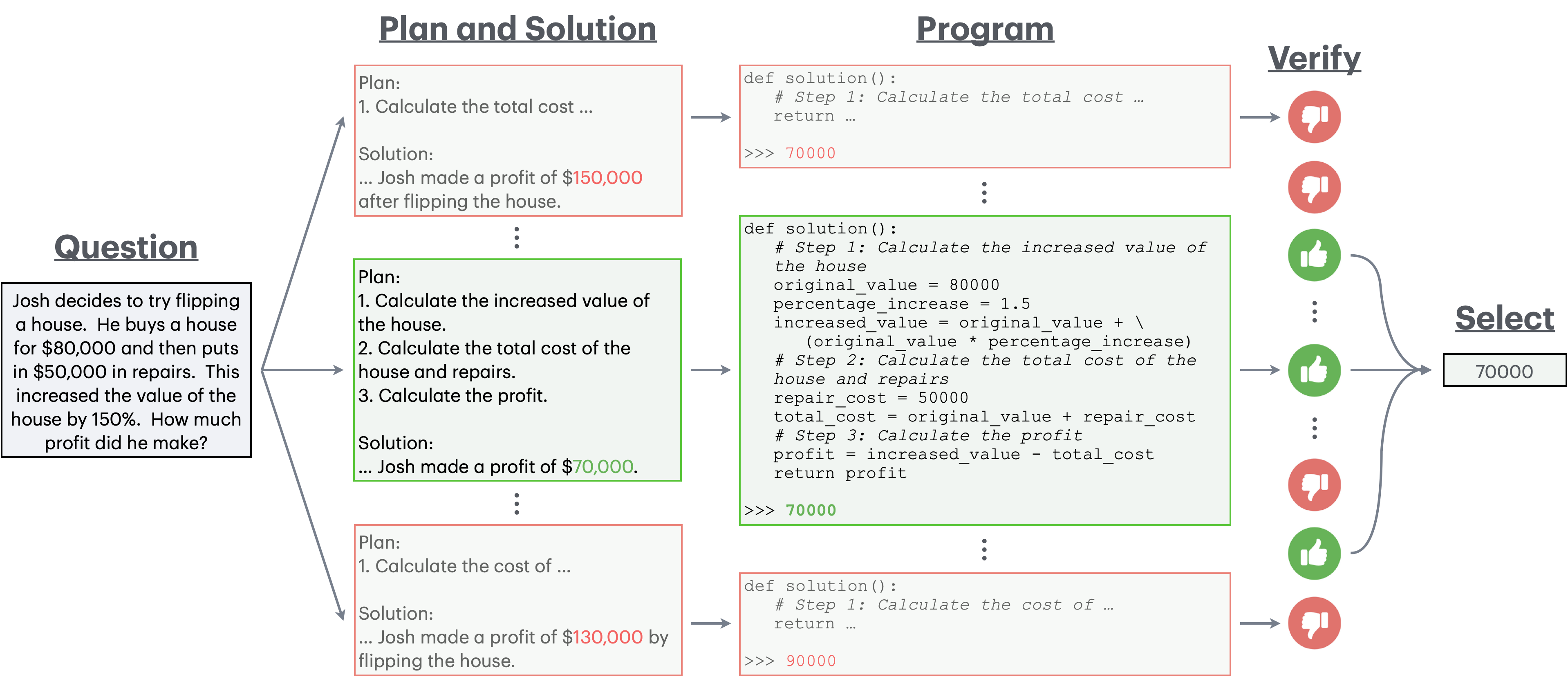}
    \caption{The \method{} Framework.}
    \label{fig:framework}
\end{figure*}

\textbf{Contributions.} We summarize the contributions of our paper as follows:
\begin{itemize}[itemsep=0pt, parsep=0pt, topsep=0pt]
  \item We propose a novel framework \method{}, that utilizes translated programs for verification, serving as a heuristic to filter out potentially incorrect reasoning paths before aggregation.
  \item Our experimental results demonstrate that \method{} significantly outperforms baseline methods for all 13 models across eight mathematical reasoning datasets.
  \item We present comprehensive ablation studies and qualitative analyses that clearly demonstrate the effectiveness of our framework.
\end{itemize}

\section{\method{}} \label{sec:method}

\paragraph{Overview.}
As illustrated in \Cref{fig:framework}, \method{} framework begins by prompting an LLM to generate a plan and solution to solve a given math word problem. Next, we prompt an LLM to translate the generated plan and solution into a Python program. We then execute the Python program and compare its output to the generated solution. If the program’s output matches the solution, it is considered valid, otherwise, the solution is filtered out. Finally, the remaining valid solutions undergo a majority voting process to determine the final answer. \method{} does not require any fine-tuning or few-shot examples. Instead, it leverages open-source LLMs with zero-shot prompting, making it highly flexible and efficient.

\paragraph{Generating plan and solution.}
The first step in our framework involves prompting an LLM to generate a natural language plan and solution to solve math word problems. For each math word problem $x$, we prompt an LLM to generate a plan and solution $s$ using the zero-shot prompt proposed by \citet{wang-etal-2023-plan}: ``\textit{Let’s first understand the problem and devise a plan to solve the problem. Then, let’s carry out the plan and solve the problem step by step}''. We sample multiple plans and solutions from the LLM to produce a set of candidate plans and solutions \{$s_{1}, ..., s_{n}$\}.

\paragraph{Translating plan and solution to Python program.}
After obtaining the set of candidate plans and solutions \{$s_{1}, ..., s_{n}$\}, we use each plan and solution $s_{i}$ individually as input to an LLM, prompting it to generate a Python program $p_{i}$ that implements the plan to solve the given math word problem. This process results in a corresponding Python program for each candidate plan and solution, yielding the pairs \{$(s_{1}, p_{1}), ..., (s_{n}, p_{n})$\}, which are then used for verification.

\paragraph{Verification and selection.}
Lastly, we execute the Python program $p_{i}$ to obtain the output $\hat{p_{i}}$, which is then used to verify the answer $\hat{s_{i}}$ from the generated solution $s_{i}$. The answer $\hat{s_{i}}$ is extracted using the answer extraction prompting method proposed by \citet{kojima2022large}: ``\textit{Therefore, the answer (arabic numerals) is}''. We consider the answer $\hat{s_{i}}$ valid if it matches the program output $\hat{p_{i}}$. After verifying all candidate plans and solutions, we perform majority voting on the remaining valid answers to determine the final answer. If no valid answer remain, majority voting is performed on the invalid answers to select the final answer.

\begin{figure*}[!ht]
    \centering
    \includegraphics[width=1\linewidth]{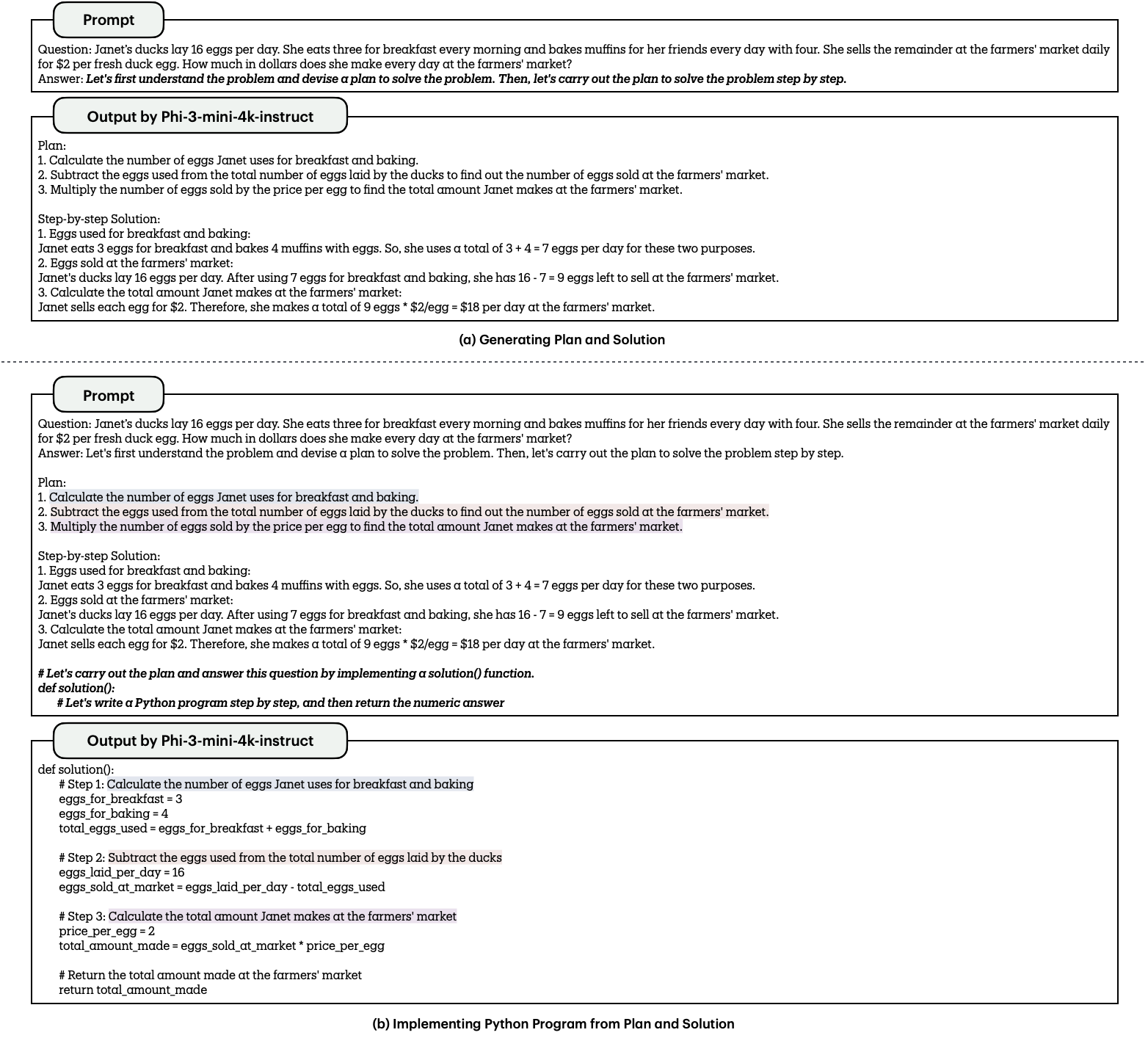}
    \caption{An actual example of the prompts used in our framework, along with the outputs generated by Phi3-mini-4k-instruct, for the GSM8K task.}
    \label{fig:framework_example}
\end{figure*}

\section{Experiments}

\subsection{Setup}

\paragraph{Models and datasets.} 
\method{} is a model-agnostic framework designed to integrate with any LLM for mathematical reasoning tasks. We evaluated 13 LLMs from various model families and sizes, ranging from 0.5B to 13B parameters. Specifically, we evaluated on Phi-3-mini-4k-instruct (3.8B) \cite{abdin2024phi3technicalreporthighly}, Llama-2-7B-chat \cite{touvron2023llama2openfoundation}, Llama-2-13B-chat, Llama-3-8B-Instruct \cite{dubey2024llama3herdmodels}, Llama-3.1-8B-Instruct, Llama-3.2-1B-Instruct, Llama-3.2-3B-Instruct, Mistral-7B-Instruct-v0.3 \cite{jiang2023mistral7b}, Gemma-2-2B-it \cite{gemmateam2024gemma2improvingopen}, Gemma-2-9B-it, Qwen2-0.5B-Instruct \cite{yang2024qwen2technicalreport}, Qwen2-1.5B-Instruct, and Qwen2-7B-Instruct. We conducted evaluations across eight mathematical reasoning datasets, including GSM8K \cite{cobbe2021gsm8k}, MATH \cite{hendrycksmath2021}, SVAMP \cite{patel-etal-2021-nlp}, ASDIV \cite{miao-etal-2020-diverse}, MultiArith \cite{roy-roth-2015-solving}, SingleEQ \cite{koncel-kedziorski-etal-2015-parsing}, SingleOP \cite{roy-etal-2015-reasoning}, and AddSub \cite{hosseini-etal-2014-learning}. Following \citet{lightman2024lets}, we use MATH-500, a subset of problems from the MATH dataset to speed up evaluation.

\begin{table*}[!ht]
    \small 
    \centering
    \resizebox{0.98\textwidth}{!}{
        \begin{tabular}{l|ccc|cc|c|c|cc|c|c|cc}
            \toprule
            \multicolumn{1}{c}{\multirow{2}{*}[-3pt]{\textbf{Method}}} & \multicolumn{3}{c}{Qwen-2} & \multicolumn{2}{c}{Gemma-2} & \multicolumn{1}{c}{Phi-3}  & \multicolumn{1}{c}{Mistral} & \multicolumn{2}{c}{Llama-2} & \multicolumn{1}{c}{Llama-3} & \multicolumn{1}{c}{Llama-3.1} & \multicolumn{2}{c}{Llama-3.2} \\
            
            \cmidrule(lr{0.5em}){2-4}\cmidrule(lr{0.5em}){5-6}\cmidrule(lr{0.5em}){7-7}\cmidrule(lr{0.5em}){8-8}\cmidrule(lr{0.5em}){9-10}\cmidrule(lr{0.5em}){11-11}\cmidrule(lr{0.5em}){12-12}\cmidrule(lr{0.5em}){13-14}
            
             & 0.5B & 1.5B & 7B & 2B & 9B & 3.8B  & 7B & 7B & 13B & 8B & 8B & 1B & 3B \\

            \midrule
            \multicolumn{14}{c}{\cellcolor[HTML]{EFEFEF}\textit{\textbf{GSM8K}}} \\
            \midrule
            Direct  & 37.00 & 61.56 & 88.02 & 67.32 & 88.63 & 85.67 & 59.06 & 30.55 & 42.84 & 79.38 & 85.67 & 48.75 & 79.23\\
            Zero-shot CoT & 32.22 & 57.71 & 88.25 & 67.63 & 89.69 & 86.35 & 58.83 & 29.80  & 41.24 & 80.74 & 87.49 & 50.72 & 81.88\\
            Zero-shot PoT & 5.23 & 2.81 & 63.61 & 50.80 & 80.89 & 84.38 & 48.52 & 8.49 & 26.99 & 73.46 & 84.38 & 42.15 & 33.06 \\
            Zero-shot PS & 25.63 & 51.02 & 85.90 & 63.00 & 89.01 & 86.50 & 58.07 & 28.35  & 39.80 & 80.36 & 86.66 & 49.05 &  81.12 \\
            \midrule
            Zero-shot CoT @maj16 & 48.85 & 71.57 & 91.66 & 72.94 & 91.05 & 91.96 & 75.66 & 38.97 & 54.44 & 87.72 & 92.49 & 66.11 & 87.87 \\
            Zero-shot PoT @maj16 & 28.43 & 37.00 & 85.67 & 59.21 & 86.43 & 91.36 & 70.96  & 27.22  & 40.86 & 84.07 & 90.37 & 58.45 & 83.32 \\
            Zero-shot PS @maj16 & 42.00 & 68.46 & 90.98 & 73.39 & 91.96 & 92.19 & 78.54 & 41.32 & 54.21 & 89.61 & 92.57 & 65.66 &88.86\\
            \midrule
            \multirow{2}{*}{\method{}} & \textbf{53.83} & \textbf{74.22} & \textbf{92.42} & \textbf{79.61} & \textbf{92.72} & \textbf{93.10} & \textbf{83.24} & \textbf{59.51} & \textbf{68.08} & \textbf{90.14} & \textbf{93.19} & \textbf{73.01} & \textbf{91.36} \\
            & \footnotesize{\highlightvalue{4.98}} & \footnotesize{\highlightvalue{2.65}} & \footnotesize{\highlightvalue{0.76}} & \footnotesize{\highlightvalue{6.22}} & \footnotesize{\highlightvalue{0.76}} & \footnotesize{\highlightvalue{0.91}} & \footnotesize{\highlightvalue{4.70}} & \footnotesize{\highlightvalue{18.19}} & \footnotesize{\highlightvalue{13.64}} & \footnotesize{\highlightvalue{0.53}} & \footnotesize{\highlightvalue{0.62}} & \footnotesize{\highlightvalue{6.90}} & \footnotesize{\highlightvalue{2.50}} \\

            \midrule
            \multicolumn{14}{c}{\cellcolor[HTML]{EFEFEF}\textit{\textbf{MATH-500}}} \\
            \midrule
            Direct  & 5.40 & 16.60 & 49.00 & 20.20 & 47.00 & 39.60 & 12.60 & 2.20 & 4.20 & 26.80 & 47.60 & 13.60 & 39.00 \\
            Zero-shot CoT & 7.20 & 18.60 & 45.20 & 19.60 & 46.40 & 41.60 & 9.60 & 2.00 & 4.40 & 26.00 & 46.60 & 13.40 &36.80 \\
            Zero-shot PS & 7.00 & 13.40 & 37.40 & 20.80 & 42.60 & 42.40 & 10.80 & 2.20 & 4.00 & 24.60 & 47.60 & 16.60 & 35.60\\
            \midrule
            Zero-shot CoT @maj16 & 11.60 & 29.40 & 64.00 & 29.20 & 53.20 & 54.00 & 18.60 & 3.20 & 8.00 & 34.80 & 62.40 & 34.60 & 54.80 \\
            Zero-shot PS @maj16 & 11.60 & 28.20 & 63.40 & 29.00 &54.80  & 51.40 & 21.20 & 3.80 & 6.20 & 36.80 & 62.80 & 38.00  & 56.40 \\
            \midrule
            \multirow{2}{*}{\method{}} & \textbf{19.80} & \textbf{34.80} & \textbf{67.20} & \textbf{37.40} & \textbf{58.60} & \textbf{57.80} & \textbf{27.00} & \textbf{8.80} & \textbf{17.40} & \textbf{44.80} & \textbf{65.60} & \textbf{44.80} & \textbf{57.40} \\
            & \footnotesize{\highlightvalue{8.20}} & \footnotesize{\highlightvalue{5.40}} & \footnotesize{\highlightvalue{3.20}} & \footnotesize{\highlightvalue{8.20}} & \footnotesize{\highlightvalue{3.80}} & \footnotesize{\highlightvalue{3.80}} & \footnotesize{\highlightvalue{5.80}} & \footnotesize{\highlightvalue{5.00}} & \footnotesize{\highlightvalue{9.40}} & \footnotesize{\highlightvalue{8.00}} & \footnotesize{\highlightvalue{2.80}} & \footnotesize{\highlightvalue{6.80}} & \footnotesize{\highlightvalue{1.00}} \\

            \midrule
            \multicolumn{14}{c}{\cellcolor[HTML]{EFEFEF}\textit{\textbf{SVAMP}}} \\
            \midrule
            Direct  & 45.70 & 65.90 & 92.40 & 72.90 & 89.70 & 91.70 & 69.30 & 61.40 & 66.20 & 83.00 & 89.10 &  70.90  & 85.70 \\
            Zero-shot CoT & 41.00 & 62.10 & 89.90 & 74.40 & 88.70 & 89.10 & 71.80 & 56.40 & 62.80 & 85.00 & 86.80  & 68.40 & 86.50\\
            Zero-shot PoT & 13.60 & 5.40 & 47.00 & 73.40 & 91.80 & 90.70 & 52.80 &  25.50 & 44.10 & 86.70 & 88.00  & 62.60 & 66.30 \\
            Zero-shot PS & 35.60 & 61.00 & 90.70 & 73.60 & 89.80 & 89.80 & 73.20 &  56.00 & 57.30 & 83.40 & 88.30 & 65.60 & 87.50 \\
            \midrule
            Zero-shot CoT @maj16 & 57.10 & 78.30 & 93.60 & 80.00 & 91.30 & 94.00 & 84.40 &  69.70 & 77.80 & 90.50 & 92.30 & 82.30 &91.90 \\
            Zero-shot PoT @maj16 & 47.80 & 39.20 & 74.50 & 79.30 & 92.50 & 94.70 & 80.50 & 59.70 & 66.00 & 90.60 & 91.50 & 76.80 & 92.00\\
            Zero-shot PS @maj16 & 56.60 & 79.60 & 93.70 & 81.70 & 91.40 & 93.60 & 84.70 &  71.30 & 77.80 & 91.20 & 92.90 & 81.50 & 92.80\\
            \midrule
            \multirow{2}{*}{\method{}} & \textbf{73.80} & \textbf{86.70} & \textbf{94.60} & \textbf{86.40} & \textbf{92.90} & \textbf{95.10} & \textbf{88.90} &  \textbf{82.70} & \textbf{86.40} & \textbf{93.90} & \textbf{94.50} & \textbf{87.70} & \textbf{93.70} \\
            & \footnotesize{\highlightvalue{16.70}} & \footnotesize{\highlightvalue{7.10}} & \footnotesize{\highlightvalue{0.90}} & \footnotesize{\highlightvalue{4.70}} & \footnotesize{\highlightvalue{0.40}} & \footnotesize{\highlightvalue{0.40}} & \footnotesize{\highlightvalue{4.20}} & \footnotesize{\highlightvalue{11.40}} & \footnotesize{\highlightvalue{8.60}} & \footnotesize{\highlightvalue{2.70}} & \footnotesize{\highlightvalue{1.60}} & \footnotesize{\highlightvalue{5.40}} & \footnotesize{\highlightvalue{0.90}} \\

            \midrule
            \multicolumn{14}{c}{\cellcolor[HTML]{EFEFEF}\textit{\textbf{ASDIV}}} \\
            \midrule
            Direct  & 60.02 & 75.43 & 91.94 & 81.30 & 91.70 & 93.32 & 73.57 & 61.31 & 67.51 & 85.26 & 89.17 & 76.91 & 88.88 \\
            Zero-shot CoT & 55.10 & 73.14 & 91.89 & 80.68 & 91.60 & 91.84 & 74.52 & 58.59 & 65.03 & 85.35 & 89.12 & 75.91 & 89.22 \\
            Zero-shot PoT & 18.46 & 6.35 & 48.09 & 71.66 & 82.20 & 83.73 & 54.20 & 32.40 & 48.71 & 78.86 & 86.74 & 67.18 & 59.49 \\
            Zero-shot PS & 45.99 & 70.04 & 91.32 & 81.82 & 91.51 & 91.98 & 74.62 & 57.63 & 62.12 & 84.40 & 88.98 & 72.47 & 87.74 \\
            \midrule
            Zero-shot CoT @maj16 & 69.18 & 82.68 & 93.56 & 85.35 & 92.56 & 94.51 & 83.92 & 66.46 & 74.86 & 89.46 & 92.27 &  84.45& 92.70 \\
            Zero-shot PoT @maj16 & 57.49 & 59.02 & 77.24 & 78.24 & 86.26 & 92.89 & 78.77 & 63.36 & 66.56 & 86.74 & 91.89 &  79.53& 91.27 \\
            Zero-shot PS @maj16 & 68.03 & 83.35 & 93.46 & 87.02 & 92.99 & 94.18 & 84.78 & 69.27 & 75.43 & 89.69 & 92.22 &  85.07& 92.75 \\
            \midrule
            \multirow{2}{*}{\method{}} & \textbf{74.90} & \textbf{84.83} & \textbf{93.76} & \textbf{87.88} & \textbf{93.35} & \textbf{94.75} & \textbf{87.02} & \textbf{76.67} & \textbf{80.49} & \textbf{89.98} & \textbf{92.83} & \textbf{86.35} & \textbf{93.03} \\
            & \footnotesize{\highlightvalue{5.72}} & \footnotesize{\highlightvalue{1.48}} & \footnotesize{\highlightvalue{0.20}} & \footnotesize{\highlightvalue{0.86}} & \footnotesize{\highlightvalue{0.36}} & \footnotesize{\highlightvalue{0.24}} & \footnotesize{\highlightvalue{2.24}} & \footnotesize{\highlightvalue{7.40}} & \footnotesize{\highlightvalue{5.06}} & \footnotesize{\highlightvalue{0.29}} & \footnotesize{\highlightvalue{0.56}} & \footnotesize{\highlightvalue{1.28}} & \footnotesize{\highlightvalue{0.28}} \\

            \midrule
            \multicolumn{14}{c}{\cellcolor[HTML]{EFEFEF}\textit{\textbf{MULTIARITH}}} \\
            \midrule
            Direct  & 79.33 & 95.83 & 97.67 & 94.00 & 98.50 & 98.33 & 78.50 & 74.50 & 82.67 & 98.50 & 97.00 & 71.83 & 96.50\\
            Zero-shot CoT &  78.17 & 93.50  & 98.67 & 94.00  &  98.17& 98.17 & 83.67 & 74.33 & 76.67 & 96.67 & 97.83 & 85.67 & 97.83\\
            Zero-shot PoT &  18.00 & 4.33  & 59.67 & 88.67 & 98.00 & 96.67 & 61.67 & 29.83 & 60.17 & 97.33 & 97.83 & 78.67 &62.33 \\
            Zero-shot PS &  62.33 & 87.50  & 97.83 & 94.33 & 98.67 & 98.50 & 83.50 & 70.67 & 70.00 & 97.83 & 97.33 & 83.67 & 97.00\\
            \midrule
            Zero-shot CoT @maj16 &  91.83 & 98.00  & \textbf{98.33} & 97.17 & 98.33 & \textbf{98.83} & 91.50 & 85.67 & 92.17 & 98.83 & 98.00 & 97.00 & \textbf{99.00} \\
            Zero-shot PoT @maj16 &  78.33 & 49.67   & 90.33 & 95.00 & 98.67 & 98.67 & 92.83 & 77.83 & 79.00 & \textbf{99.00} & 99.00 & 95.83 & \textbf{99.00}\\
            Zero-shot PS @maj16 &  91.33 & 98.33   & 98.17 & 97.00 & 98.67 & \textbf{98.83} & 94.33 & 90.33 & 92.17 & 98.17 & 98.50 & \textbf{97.83}  & 98.50\\
            \midrule
            \multirow{2}{*}{\method{}} &  \textbf{95.50} & \textbf{99.33}   & \textbf{98.33} & \textbf{98.33} & \textbf{98.83} & \textbf{98.83} & \textbf{95.67} & \textbf{97.67} & \textbf{96.17} & \textbf{99.00} & \textbf{99.17} & \textbf{97.83} & \textbf{99.00} \\

            & \footnotesize{\highlightvalue{3.67}} & \footnotesize{\highlightvalue{1.00}} & \footnotesize{\highlightvalue{0.00}} & \footnotesize{\highlightvalue{1.16}} & \footnotesize{\highlightvalue{0.16}} & \footnotesize{\highlightvalue{0.00}} & \footnotesize{\highlightvalue{1.34}} & \footnotesize{\highlightvalue{7.34}} & \footnotesize{\highlightvalue{4.00}} & \footnotesize{\highlightvalue{0.00}} & \footnotesize{\highlightvalue{0.17}} & \footnotesize{\highlightvalue{0.00}} & \footnotesize{\highlightvalue{0.00}} \\

            \midrule
            \multicolumn{14}{c}{\cellcolor[HTML]{EFEFEF}\textit{\textbf{SINGLEEQ}}} \\
            \midrule
            Direct  &  72.44 & 88.19  &95.47& 91.14 & 94.88 & 96.06 & 85.24 & 79.92 & 82.68 & 93.50 & 93.31 & 89.96 & 94.69\\
            Zero-shot CoT &  67.72 & 85.24  & 97.44 & 92.32 & 96.65 & 98.23 & 83.86 & 78.54 & 83.07 & 93.70 & 94.49 & 87.40 & 95.47\\
            Zero-shot PoT &  25.00 & 5.91   & 49.61 & 94.49 & 98.23 & 98.03 & 69.49 & 41.34 & 69.69 & 94.49 & 96.26 & 84.06 & 62.40\\
            Zero-shot PS &  62.80 & 81.69  & 96.65 & 92.13 & 97.24 & 98.03 & 85.83 & 75.00 & 76.57 & 91.14 & 95.28 & 84.45 & 93.90\\
            \midrule
            Zero-shot CoT @maj16 &  81.50 & 90.75  & 96.85 & 94.49 & 97.24 & 98.82 & 89.37 & 83.27 & 92.13 & 96.85 & 97.24 & 94.09 & 97.83\\
            Zero-shot PoT @maj16 &  80.31 & 71.06  &  79.33 & 96.06 & 98.82 & 99.02 & \textbf{94.69} & 86.22 & 88.78 & 98.03 & 98.43 & 94.49 & 97.64\\
            Zero-shot PS @maj16 &  80.71 & 93.11  & 97.44 & 95.08 & 98.03 & 98.23 & 91.14 & 85.24 & 91.73 & 97.24 & 97.83 & 94.88 & 97.24\\
            \midrule
            \multirow{2}{*}{\method{}} &  \textbf{92.13} & \textbf{96.80}  & \textbf{98.82} & \textbf{97.64} & \textbf{99.21} & \textbf{99.21} & \textbf{94.69} & \textbf{92.91} & \textbf{97.24} & \textbf{98.23} & \textbf{98.62} &  \textbf{96.06} & \textbf{98.82} \\

            &\footnotesize{\highlightvalue{10.63}} & \footnotesize{\highlightvalue{3.69}} & \footnotesize{\highlightvalue{1.38}} & \footnotesize{\highlightvalue{1.58}} & \footnotesize{\highlightvalue{0.39}} & \footnotesize{\highlightvalue{0.19}} & \footnotesize{\highlightvalue{0.00}} & \footnotesize{\highlightvalue{6.69}} & \footnotesize{\highlightvalue{5.11}} & \footnotesize{\highlightvalue{0.20}} & \footnotesize{\highlightvalue{0.19}} & \footnotesize{\highlightvalue{1.18}} & \footnotesize{\highlightvalue{0.99}} \\

            \midrule
            \multicolumn{14}{c}{\cellcolor[HTML]{EFEFEF}\textit{\textbf{SINGLEOP}}} \\
            \midrule
            Direct  &  77.76 & 90.75  & 96.09 & 91.28 & 95.37 & 96.8 & 90.04 & 83.45 & 88.79 & 93.95 & 95.73 & 92.35 & 96.26\\
            Zero-shot CoT &  6.33 & 89.32   & 96.26 & 91.64 & 95.91 & 96.26 & 90.93 & 85.41 & 85.94 & 94.13 & 96.09 & 88.43 & 95.91\\
            Zero-shot PoT &  29.54 & 10.5  & 44.84 & 90.57 &95.73 & 95.55 & 72.78 & 51.25 & 68.51 & 93.24 & 95.91 & 84.70 & 74.02\\
            Zero-shot PS &  66.19 & 88.26  & 96.26 & 93.42 & 96.26 & 96.80 & 90.75 & 81.67 & 81.32 & 91.10 & 95.55 & 86.83 & 94.13 \\
            \midrule
            Zero-shot CoT @maj16 &  87.37 & 92.88  & 96.26 & 94.13 & 96.09 & 96.80 & 94.66 & 88.79 & 92.70 & 96.62 & 96.80 & 93.77 & \textbf{96.80}\\
            Zero-shot PoT @maj16 &  84.52 & 77.22  & 80.60 & 92.88 & 95.73 & 96.80 & 93.59 & 89.86 & 85.77 & 95.37 & 96.44 & 91.99 & 96.44\\
            Zero-shot PS @maj16 &  88.08 & 93.59  & 96.26 & 94.66 & \textbf{96.62} & 97.15 & 94.84 & 91.10 & 91.28 & 96.44 & 96.62 & 93.06 & 96.62\\
            \midrule
            \multirow{2}{*}{\method{}}&  \textbf{92.53} & \textbf{94.66}   & \textbf{96.98} & \textbf{94.84} & \textbf{96.62} & \textbf{97.51} & \textbf{95.02} & \textbf{94.66} & \textbf{93.24} & \textbf{97.16} & \textbf{96.98} & \textbf{94.66} & \textbf{96.80} \\
            & \footnotesize{\highlightvalue{4.45}} & \footnotesize{\highlightvalue{1.07}} & \footnotesize{\highlightvalue{0.72}} & \footnotesize{\highlightvalue{0.18}} & \footnotesize{\highlightvalue{0.00}} & \footnotesize{\highlightvalue{0.36}} & \footnotesize{\highlightvalue{0.18}} & \footnotesize{\highlightvalue{3.56}} & \footnotesize{\highlightvalue{0.54}} & \footnotesize{\highlightvalue{0.54}} & \footnotesize{\highlightvalue{0.18}} & \footnotesize{\highlightvalue{0.89}} & \footnotesize{\highlightvalue{0.00}} \\

            \midrule
            \multicolumn{14}{c}{\cellcolor[HTML]{EFEFEF}\textit{\textbf{ADDSUB}}} \\
            \midrule
            Direct  &  61.27 & 80.51  & 92.91 & 84.05 & 90.38 & 93.16 & 81.01 & 73.16 & 82.53 & 88.10 & 89.62 & 84.56 & 90.38\\
            Zero-shot CoT &  58.73 & 79.75  & 94.94 & 86.33 & 94.94 & 95.95 & 82.28 & 73.67 & 81.01 & 89.37 & 92.66 & 84.56 & 93.42\\
            Zero-shot PoT &  49.62 & 7.34   & 28.61 & 53.92 & 95.95 & 95.7 & 70.38 & 42.53 & 65.57 & 93.67 & 95.44 & 78.73 & 88.86\\
            Zero-shot PS &  51.39 & 77.47  & 92.91 & 89.11 & 95.19 & 94.43 & 83.29 & 73.42 & 75.19 & 85.32 & 92.66 & 77.97 & 90.89\\
            \midrule
            Zero-shot CoT @maj16 &  70.89 & 86.58  & 95.44 & 90.89 & 94.43 & 96.2 & 85.82 & 80.51 & 86.08 & 93.42 & 95.19 & 89.62 & \textbf{95.95} \\
            Zero-shot PoT @maj16 &  71.39 & 64.81   & 63.54 & 90.89 & 95.95 & 96.46 & 91.14 & 81.52 & 85.06 & \textbf{95.19} & 95.95 & 82.53 & \textbf{95.95} \\
            Zero-shot PS @maj16 &  71.14 & 85.82  & 94.94 & 91.14  & 95.7 & 96.46 & 88.35 & 83.29 & 87.34 & 93.16 & 94.43 & 90.13 & 94.68\\
            \midrule
            \multirow{2}{*}{\method{}} & \textbf{85.06} & \textbf{92.15}  & \textbf{96.46} & \textbf{92.41} & \textbf{96.20} & \textbf{96.96} & \textbf{91.65} & \textbf{89.37} & \textbf{93.16} & \textbf{95.19} & \textbf{96.20} &  \textbf{92.41} & \textbf{95.95} \\

            & \footnotesize{\highlightvalue{13.67}} & \footnotesize{\highlightvalue{5.57}} & \footnotesize{\highlightvalue{1.02}} & \footnotesize{\highlightvalue{1.27}} & \footnotesize{\highlightvalue{0.25}} & \footnotesize{\highlightvalue{0.50}} & \footnotesize{\highlightvalue{0.51}} & \footnotesize{\highlightvalue{6.08}} & \footnotesize{\highlightvalue{5.82}} & \footnotesize{\highlightvalue{0.00}} & \footnotesize{\highlightvalue{0.25}} & \footnotesize{\highlightvalue{2.28}} & \footnotesize{\highlightvalue{0.00}} \\

            \bottomrule
        \end{tabular}
     }
    \caption{Main results showing the comparison of \method{} and baseline methods across 13 LLMs, ranging in size from 0.5B to 13B, on eight mathematical reasoning datasets.}
    \label{tab:mainresults}
\end{table*}

\paragraph{Implementation details.}
In \method{}, we sample a total of 16 plans and solutions using a temperature of 0.7. To extract the answer from the generated solutions, we rely on Phi-3-mini-4k-instruct. We also standardize the use of Phi-3-mini-4k-instruct to translate the plan and solution into Python programs for verification. With only 3.8 billion parameters, Phi-3-mini-4k-instruct offers more efficient inference. Notably, when Phi-3-mini-4k-instruct is being evaluated, it performs self-verification by generating not only the plan and solution but also translating the plan and solution to a Python program to verify the results. During the selection phase, we perform majority voting on the remaining valid answer. If no valid answer remain, we perform majority voting over all invalid answers. For the more challenging MATH dataset, we use GPT-4o (2024-02-15-preview) to translate the plan and solution into Python code. Additional details about the MATH dataset setup can be found in \Cref{app:math_implementation}.

\paragraph{Baselines.} 
We compare \method{} with two decoding strategies: single sampling (greedy decoding) and multiple sampling (self-consistency decoding) \cite{wang2023selfconsistency}. For the comparison, we use three different prompting techniques: zero-shot CoT \cite{kojima2022large}, zero-shot PoT \cite{chen2022program}, and zero-shot PS \cite{wang-etal-2023-plan}. Performance is evaluated based on accuracy, where a problem is considered correct only if the final answer matches the ground truth.

\begin{figure*}[t]
    \centering
    \includegraphics[width=1\linewidth]{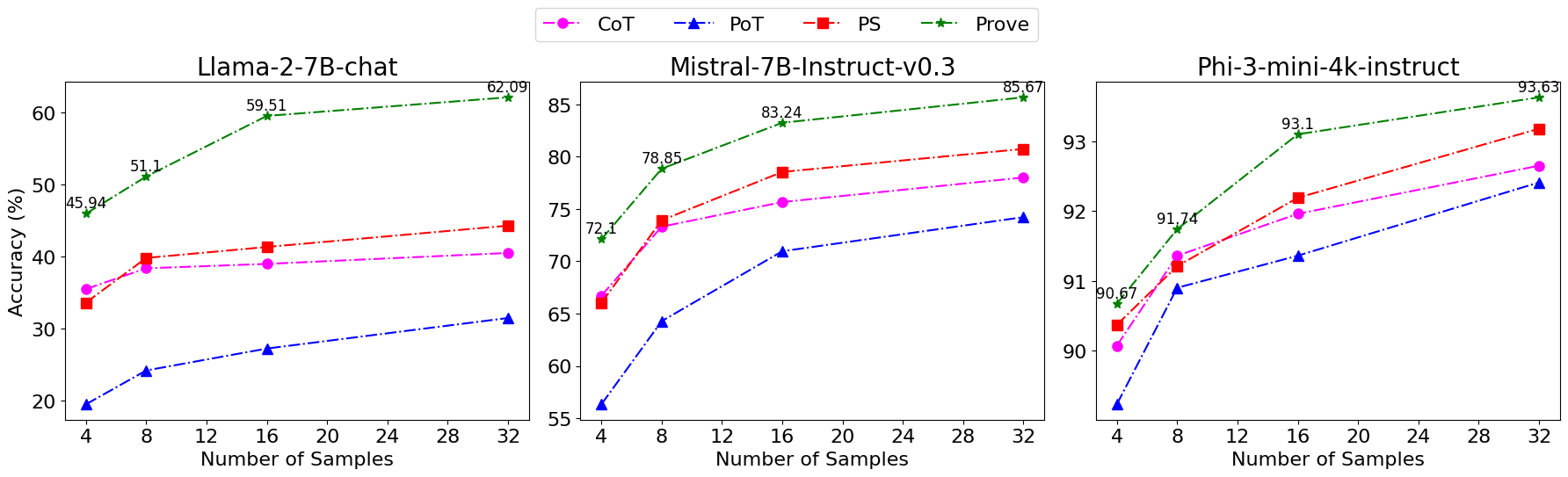}
    \caption{Performance comparison of \method{} and baseline methods across various LLMs evaluated on GSM8K with different numbers of samples.}
    \label{fig:very_plans}
\end{figure*}

\subsection{Main Results}
We report the main results of our experiments in \Cref{tab:mainresults}. We highlight several key observations:

\paragraph{\method{} improves over baselines consistently on all evaluated LLMs across eight datasets.} 
Our evaluation of \method{} across various model families and sizes demonstrates consistent performance improvements over strong baseline methods. For example, on the GSM8K dataset, Mistral-7B-Instruct-v0.3 achieves an accuracy of 83.24\% with \method{}, a 4.7\% increase compared to 78.54\% achieved by the Zero-shot PS @maj16 baseline. Similarly, on the SVAMP dataset, Mistral-7B-Instruct-v0.3 attains an accuracy of 88.90\%, surpassing the best baseline by 4.2\%. Additionally, Mistral-7B-Instruct-v0.3 delivers performance gains of up to 3.39\% across other datasets, including ASDIV, MultiArith, SingleEq, SingleOp, and AddSub. These improvements are consistently observed across all 13 LLMs we evaluated.

\paragraph{\method{} provides larger performance gains for smaller models.} 

In our comparison of various model families including Qwen-2, Gemma-2, Llama-2, and Llama-3.2, we observed that smaller models tend to show greater performance improvements than their larger counterparts. For instance, on the GSM8K dataset, the smaller Llama-2-7B-chat model achieved an 18.19\% improvement over the baseline, whereas the larger Llama-2-13B-chat model saw only a 13.87\% gain. Similarly, the Gemma-2-2B-it model improved by 6.22\%, while the larger Gemma-2-9B-it model achieved just a 0.76\% increase. A similar trend is also evident in the Qwen-2 models: Qwen2-0.5B-Instruct outperformed its baseline by 4.98\%, Qwen2-1.5B-Instruct improved by 2.65\%, and the larger Qwen2-7B-Instruct managed only a 0.76\% gain. Likewise, Llama-3.2-1B-Instruct demonstrated a 7.35\% improvement, whereas its larger variant, Llama-3.2-3B-Instruct, only achieved a 2.5\% increase. These results suggest that smaller models benefit more substantially from \method{}, likely because they are more susceptible to errors, such as miscalculations, making filtering mechanisms particularly effective for them.

\paragraph{Newer models closing the gap with larger ones.}
Recent Llama models, such as Llama-3.2, demonstrate significant progress in narrowing the performance gap with older, larger models like Llama-2 and Llama-3, despite being smaller in size. For instance, using \method{} on datasets like GSM8K, Llama-3.2-1B-Instruct achieves 73.01\% accuracy, surpassing Llama-2-13B-chat by 4.93\%. Similarly, Llama-3.2-3B-Instruct outperforms Llama-3-8B-Instruct with a score of 91.36\% compared to 90.14\%. On datasets like ASDIV, Llama-3.2-1B-Instruct and Llama-3.2-3B-Instruct further extend this lead, showcasing advancements in architecture and training techniques that enhance the performance of smaller models \cite{meta2024}.

\paragraph{Improvement on challenging mathematical dataset.} We also evaluate the effectiveness of \method{} on MATH-500. Our results show that \method{} consistently improves results across all models with the most improvement of 9.4\% for Llama-2-13B-chat. Even for smaller models such as Qwen2-0.5B-Instruct and Llama-3.2-1B-Instruct, we can see a noticeable performance gain of 8.20\% and 6.80\% respectively. These findings suggest that, even for more challenging datasets, models are capable of correctly answering questions. Furthermore, leveraging better heuristics to aggregate reasoning paths can lead to substantial performance improvements.

\subsection{Ablation Study}

\paragraph{Impact of the number of plans.} 
In \Cref{fig:very_plans}, we compare the accuracy of CoT, PoT, and PS prompting methods with self-consistency and \method{} across different numbers of sampled solutions on GSM8K. As the number of sampled solutions increases from 4 to 32, \method{} consistently outperforms the baseline methods, demonstrating its effectiveness. Notably, with just 4 samples, the Llama-2-7b-chat and Mistral-7B-Instruct-v0.3 models show significant accuracy gains of 10.46\% and 5.39\%, respectively. With 32 samples, Llama-2-7b-chat and Mistral-7B-Instruct-v0.3 models have a gain of 17.81\% and 4.93\% respectively.

\begin{table}[t]
\centering
\resizebox{1\linewidth}{!}{
\footnotesize
\begin{tabular}{lcc}
\toprule
Model & Translation Model & Accuracy (\%) \\ 
\midrule
\multirow{3}{*}{Llama-3.2-1B-Instruct}   & Phi-3-mini-4k-instruct     &      73.01     \\ 
                            & Llama-3-8B-Instruct      &      73.54     \\ 
                            &   Llama-3.1-8B-Instruct       &   74.53       \\ 
                            
\midrule
\multirow{3}{*}{Llama-2-7B-chat}   & Phi-3-mini-4k-instruct     &      59.51      \\ 
                            & Llama-3-8B-Instruct      &      58.15      \\ 
                            &   Llama-3.1-8B-Instruct       &     61.94       \\ 
\midrule
\multirow{3}{*}{Mistral-7B-Instruct-v0.3}   & Phi-3-mini-4k-instruct     &      83.24      \\ 
                            & Llama-3-8B-Instruct      &      80.29     \\ 
                            &   Llama-3.1-8B-Instruct       &   83.32       \\                
\bottomrule
\end{tabular}}
\caption{Performance of \method{} on GSM8K, evaluated across various translation models and LLMs.}
\label{fig:vary_verifier}
\end{table}

\paragraph{Impact of translation model.}
We study the impact of model selection for translating plans and solutions into Python programs on GSM8K. In our main experiments, the default translation model is Phi-3-mini-4k-instruct. Additionally, we evaluate Llama-3-8B-Instruct and Llama-3.1-8B-Instruct as alternative translation models.
\Cref{fig:vary_verifier} illustrates the performance differences across these translation models. Notably, Llama-3.1-8B-Instruct, which has more parameters than Phi-3-mini-4k-instruct, results in slight performance improvements across all tested LLMs. Specifically, it achieves performance gains of 1.52\%, 2.43\%, and 0.08\% for Llama-3.2-1B-Instruct, Llama-2-7B-chat, and Mistral-7B-Instruct-v0.3, respectively.

\begin{figure*}[t]
    \centering
    \includegraphics[width=1\linewidth]{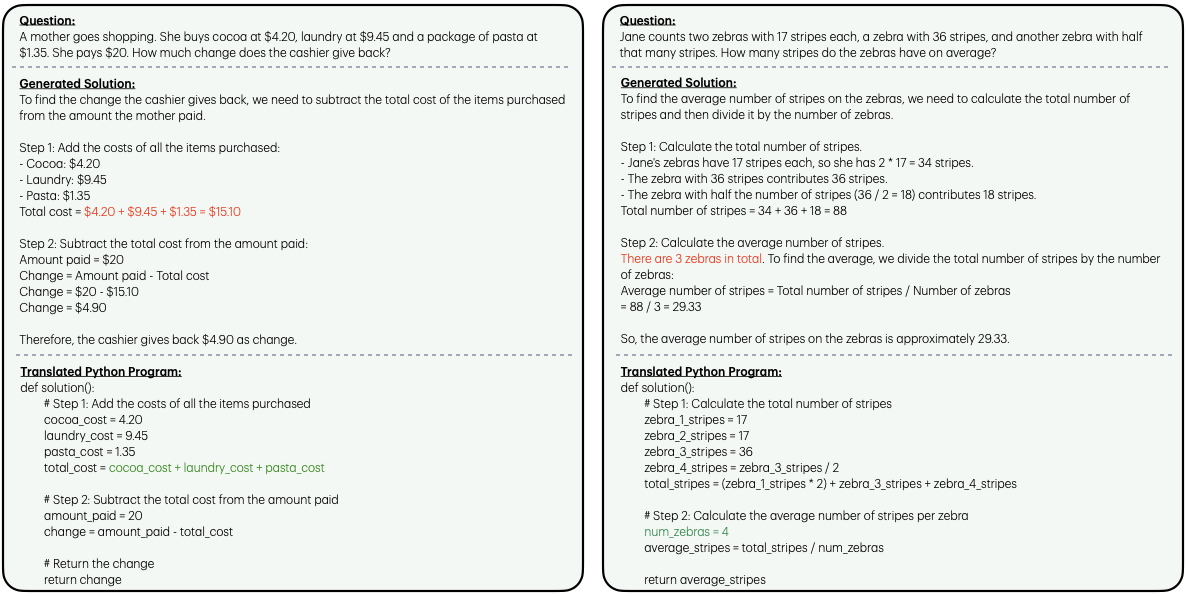}
    \caption{Case study examples where \method{} gets correct when vanilla majority voting fails on GSM8K type question. We show how the translated program can correctly filter out generated solutions that have errors in them. Left: highlighted in red shows a calculation error. Right: highlighted in red shows semantic understanding error.}
    \label{fig:case_study}
\end{figure*}

\begin{table}[t]
\centering
\resizebox{1\linewidth}{!}{
\footnotesize
\begin{tabular}{lcc}
\toprule
Model & Prompting Method & Accuracy (\%) \\ 
\midrule
\multirow{3}{*}{Llama-3.2-1B-Instruct}   & Direct    &      72.59     \\ 
                            & CoT      &      72.89     \\ 
                            &   PS       &     73.01     \\ 
                            
\midrule
\multirow{3}{*}{Llama-2-7B-chat}   & Direct    &      59.06     \\ 
                            & CoT      &      59.29     \\ 
                            &   PS       &     59.51     \\ 
\midrule
\multirow{3}{*}{Mistral-7B-Instruct-v0.3}   & Direct    &      82.34     \\ 
                            & CoT      &      82.79     \\ 
                            &   PS       &      83.24    \\                
\bottomrule
\end{tabular}
}
\caption{Performance of \method{} on GSM8K, with different prompting strategies across various LLMs.}
\label{fig:vary_prompting}
\end{table}

\paragraph{Impact of Plan and Solve (PS) prompting.} 
We investigated the impact of PS prompting on our framework by comparing its effectiveness against both direct prompting and CoT prompting in the first stage of \method{}. We evaluated three different models, namely Llama-3.2-1B-Instruct, Llama-2-7B-Chat, and Mistral-7B-Instruct-v0.3, on GSM8K. As shown in \Cref{fig:vary_prompting}, the overall performance differences between the prompting methods are relatively small. 
These findings highlights the robustness of \method{} and its ability to maintain strong performance regardless of the prompting method used.

\subsection{Analyses}

\paragraph{\method{} gets correct when the vanilla majority vote fails.}
In \Cref{fig:case_study}, we present two case study examples where \method{} successfully filter out generated solutions with errors. These two examples are examples that majority voting fails to solve due to the correct answer not being the most common answer. In contrast, \method{} effectively identifies and filter out incorrect reasoning paths, ensuring the correct answer becomes the majority. In \Cref{fig:case_study} (left), the model incorrectly calculates ``$\$4.20 + \$9.45 + \$1.35 = \$15.10$'', when the correct sum should be ``$\$15.00$''. However, using our \method{} framework, the translated program detects this miscalculation and filters out the erroneous solution. Similarly in \Cref{fig:case_study} (right), the model makes a semantic error by incorrectly determining the total number of zebras as 3 instead of 4. Once again, our translated program identifies the correct value and filters out the flawed solution. These examples demonstrate how leveraging programs as verifiers allows us to effectively filter out reasoning paths containing errors, such as miscalculations and semantic misunderstandings.

\begin{table}[t]
\centering
\footnotesize
\resizebox{1\linewidth}{!}{
\begin{tabular}{lccc}
\toprule
Method & Calculation & Missing & Semantic \\ 
\midrule
Zero-shot PS @maj16    & 20\%   &      12\%  & 68\%   \\ 
\method{}       & 0\%     &      18\%   & 82\%    \\ 
\bottomrule
\end{tabular}
}
\caption{Distribution of error types (calculation error, missing step error, semantic understanding error) for 50 randomly sampled examples from GSM8K, where both PS and \method{} got incorrect.}
\label{fig:manual_analysis}
\end{table}

\paragraph{Error analysis.} 
We conducted a manual analysis of 50 randomly sampled examples from GSM8K, where both PS and \method{} got incorrect, using Llama-3.2-1B-Instruct. Following \cite{wang-etal-2023-plan}, we categorized the errors into three types: calculation errors, missing step errors, and semantic understanding errors. The results of this analysis (\Cref{fig:manual_analysis}) show that \method{} achieves a 0\% calculation error rate, significantly lower than PS, which had a 20\% calculation error rate. This demonstrates that \method{} effectively reduces calculation errors, as shown in the case study example in \Cref{fig:case_study} (left).

\paragraph{Analysis on Python program translation.}
To assess the program translation error rate, we randomly selected 5,000 correct and incorrect PS solutions and translated them into Python programs using the default translation model used in our main experiments. 
The results show that the translation model achieves a high True Positive Rate (TPR) and a low False Negative Rate (FNR), indicating it's effectiveness in translating PS solutions into Python programs. 
Although the False Positive Rate (FPR) and True Negative Rate (TNR) are moderate, indicating that some incorrect solutions are not filtered out, the majority of incorrect PS solutions are still successfully filtered out. 
These findings provide strong evidence of the effectiveness of \method{} in mathematical reasoning.

\begin{table}[t]
\centering
\centering
\resizebox{1\linewidth}{!}{
\footnotesize
\begin{tabular}{c|c|c}
\hline
 & \textbf{Actual Positive:} & \textbf{Actual Negative} \\ 
 & Correct PS Solution & Wrong PS Solution \\ \hline
\textbf{Predicted Positive:} & \multirow{2}{*}{True Positives (TPR): 97.3\%} & \multirow{2}{*}{False Positives (FPR): 33.9\%} \\ 
Program Match &  & \\ \hline
\textbf{Predicted Negative:} & \multirow{2}{*}{False Negatives (FNR): 2.7\%} & \multirow{2}{*}{True Negatives (TNR): 66.1\%} \\ 
Program Mismatch &  &  \\ \hline
\end{tabular}
}
\caption{Confusion Matrix for program translation.}
\label{tab:confusion-matrix}
\end{table}

\section{Related Works}

\paragraph{Reasoning with Large Language Models.}
Recent advances in LLMs have demonstrated remarkable capabilities in tackling complex reasoning tasks. Prior research highlights that step-by-step reasoning prompts, such as CoT \citep{kojima2022large} and PS \cite{wang-etal-2023-plan}, improve performance compared to directly generating answers. Other techniques, such as multi-step decoding \cite{yao2023tree}, explore diverse reasoning paths, while methods like multi-sample reasoning (e.g., majority voting \cite{wang2023selfconsistency}) aggregate solutions to enhance robustness. However, these approaches still heavily depend on the LLM's generative outputs, which are prone to hallucination, particularly in smaller models. Our method, \method{}, complements these by introducing a verification step that uses Python programs to filter out potentially flawed reasoning paths, resulting in more accurate solution aggregation.

\paragraph{Large Language Models as verifiers.}
Using language models to evaluate model generations has been a long-standing idea \cite{kushman-etal-2014-learning,roy-roth-2015-solving,shen-etal-2021-generate-rank}. A common approach involves training a separate verification model to assess the correctness of generated outputs \cite{cobbe2021gsm8k}. Other techniques, such as step-by-step verification \cite{lightman2024lets} and ranking multiple reasoning paths to choose the most accurate one \cite{weng-etal-2023-large}, show promise in reducing errors but often rely heavily on the inherent reasoning abilities of the model, limiting their effectiveness, especially in smaller models.
While program-based verification approaches have been proposed \cite{zhou2024solving,han2024veritymath,zhou2024dont}, they tend to be more complex, often requiring few-shot prompting and more capable models. 
In contrast, \method{} does not require few-shot exemplars, and is easily adaptable to smaller, open-source LLMs.

\paragraph{Discussion about contemporary work.}
Recently, a contemporary study \citep{liang2024improvingllmreasoningscaling} introduced a method called CoTnPoT, which combines language and code-based answers during solution verification for mathematical reasoning. While their approach shares similarities with ours, their evaluation is limited to models with sizes of 7B and above. In contrast, our study also evaluates models ranging from 0.5B to 7B. Additionally, their work focuses exclusively on CoT prompting, whereas we also explore and analyze direct and PS prompting techniques.

\section{Conclusion}
In this paper, we demonstrate that using translated programs for verification can effectively filter out low-quality reasoning paths, which may include calculation or semantic understanding errors. Our approach, \method{}, is model-agnostic and does not require fine-tuning or few-shot exemplars for prompting. \method{} consistently outperforms baseline methods across 13 LLMs and eight mathematical reasoning datasets.

\section{Limitations}
One limitation of our approach is the need for a model that can translate the plan and solution into Python programs. For simpler mathematical reasoning datasets like GSM8K and SVAMP, we can use Phi3-mini-4k-instruct as the translation model. However, for more challenging datasets such as MATH, we require a stronger and more capable model such as GPT-4o as the translation model.

\bibliography{custom}

\appendix

\onecolumn
\section{Implementation Details for MATH} \label{app:math_implementation}

\begin{figure*}[h]
    \centering
    \includegraphics[width=1.0\linewidth]{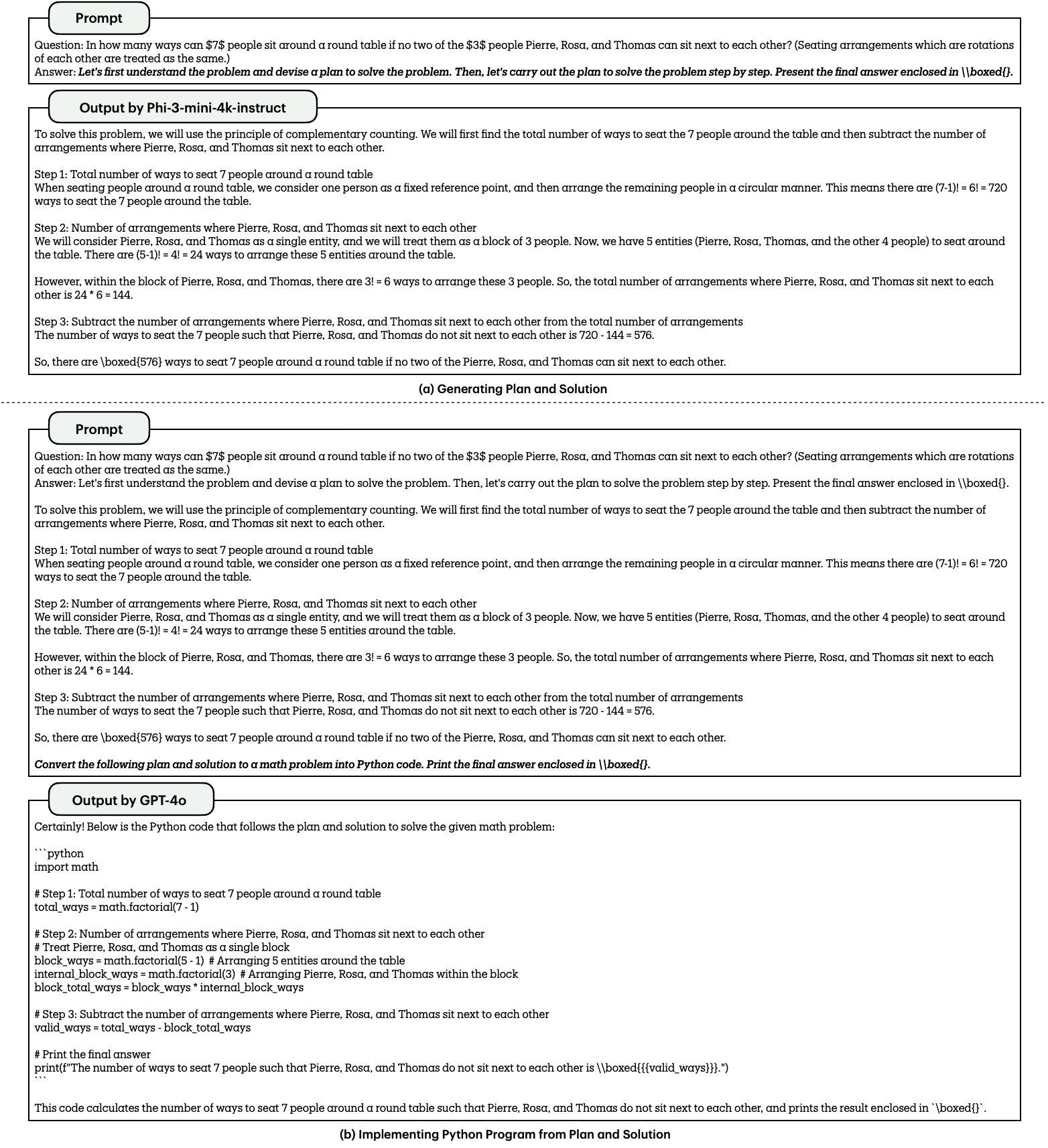}
    \caption{An actual example of prompts used for MATH dataset, along with the outputs generated by Phi3-mini-4k-instruct and GPT-4o to translate the plan and solution into Python code.}
    \label{fig:framework_example_math}
\end{figure*}

The overall framework for the MATH dataset closely follows the structure outlined in \Cref{sec:method}, with minor adjustments in the prompts to accommodate answers in LaTeX format. \Cref{fig:framework_example_math} provides an actual example of these prompts, along with the outputs generated by Phi3-mini-4k-instruct.

\paragraph{Generating plan and solution.}
We prompt an LLM to generate a plan and solution using the zero-shot prompt proposed by \citet{wang-etal-2023-plan}: ``\textit{Let’s first understand the problem and devise a plan to solve the problem. Then, let’s carry out the plan and solve the problem step by step}''. Additionally, we extend this prompt by adding ``\textit{Present the final answer enclosed in \textbackslash\textbackslash boxed\{\}.}’’, instructing the model to present the final answer within \textit{\textbackslash\textbackslash boxed\{\}} to facilitate easy extraction of the answer.

\paragraph{Translating plan and solution to Python programs.}
We use each plan and solution individually as input to GPT-4o (2024-02-15-preview), prompting it to generate a Python code block that implements the plan to solve the math word problem and printing the final answer in \textit{\textbackslash\textbackslash boxed\{\}}. We append ``\textit{Convert the following plan and solution to a math problem into Python code. Print the final answer enclosed in \textbackslash\textbackslash boxed\{\}.}'' to the prompt for generation.

\paragraph{Verification and selection.}
Finally, we execute the Python code block and retrieve the output answer by extracting the value enclosed in \textit{\textbackslash\textbackslash boxed\{\}} from the standard output produced by the \textit{print()} statement. This output answer is then compared to the generated solution for verification. If the solution matches the output answer, we consider the solution valid. After verifying all candidate plans and solutions, we conduct majority voting among the valid answers to determine the final answer. If no valid answers remain, we perform majority voting across all invalid answers instead.

\section{Case Study on MATH Dataset} \label{app:math_case_study}

We an example where \method{} succeeds in answering the question correctly, unlike the baseline methods. In this example, the problem requires differentiation, but Phi-3-mini-4k-instruct performs the differentiation incorrectly, leading to an incorrect answer. In contrast, \method{} avoids such errors by offloading the differentiation calculation to the \textit{sympy} library, allowing it to filter out flawed reasoning paths effectively.

\newpage
\newpage

\newpage
\begin{tcolorbox}[colback=gray!10, colframe=black, width=\textwidth]

\tiny
\tcbsubtitle{Question from MATH dataset}
Find the minimum value of \[\frac{(x + 5)(x + 2)}{x + 1}\]for $x > 0.$

\tcbsubtitle{Generated Solution by Phi-3-mini-4k-instruct}

We want to find the minimum value of the expression \[\frac{(x + 5)(x + 2)}{x + 1}\] for $x > 0$.

One approach is to use calculus. We can find the critical points of the function by taking the derivative and setting it equal to zero.

The derivative of the function is:
\[\textcolor{red}{\frac{d}{dx}\left(\frac{(x + 5)(x + 2)}{x + 1}\right) = \frac{(2x + 7)(x + 1) - (x + 5)(1)}{(x + 1)^2} = \frac{x^2 + 8x + 7}{(x + 1)^2}.}\]

Setting the derivative equal to zero, we get:
\[x^2 + 8x + 7 = 0.\]

This quadratic equation does not have real solutions, which means there are no critical points.

Since the function is continuous and differentiable for $x > 0$, and there are no critical points, we can check the behavior of the function as $x$ approaches infinity and zero.

As $x$ approaches infinity, the function approaches the limit:
\[\lim_{x \to \infty} \frac{(x + 5)(x + 2)}{x + 1} = \lim_{x \to \infty} \frac{x^2 + 7x + 10}{x + 1} = \infty.\]

As $x$ approaches zero from the right (since $x > 0$), the function approaches:
\[\lim_{x \to 0^+} \frac{(x + 5)(x + 2)}{x + 1} = \lim_{x \to 0^+} \frac{x^2 + 7x + 10}{x + 1} = 10.\]

Since the function approaches infinity as $x$ approaches infinity and approaches 10 as $x$ approaches zero from the right, we can conclude that the minimum value of the function is $\boxed{10}$.

\tcbsubtitle{Translated Python Program by GPT-4o}

Certainly! Below is the Python code that follows the plan and solution to find the minimum value of the given expression:
\begin{lstlisting}[language=Python, basicstyle=\tiny]

```python
import sympy as sp

# Define the variable
x = sp.symbols('x', positive=True)

# Define the function
y = (x + 5) * (x + 2) / (x + 1)

# Simplify the function
y_simplified = sp.simplify(y)

# Find the derivative
y_prime = sp.diff(y_simplified, x)

# Solve for critical points
critical_points = sp.solve(y_prime, x)

# Check the second derivative to confirm minimum
y_double_prime = sp.diff(y_prime, x)

# Evaluate the second derivative at critical points
min_value = None
for point in critical_points:
    if y_double_prime.subs(x, point) > 0:
        min_value = y_simplified.subs(x, point)
        break

# Print the minimum value
print(f"The minimum value is: \\boxed{{{min_value}}}")
```
\end{lstlisting}
This code uses the `sympy` library to perform symbolic mathematics, including differentiation and solving equations. It finds the critical points, checks the second derivative to confirm a minimum, and evaluates the function at the critical point to find the minimum value.

\end{tcolorbox}
\newpage

\end{document}